\documentclass[letterpaper]{article} 

\usepackage{aaai2026}
\usepackage{times}  
\usepackage{helvet}  
\usepackage{courier}  
\usepackage[hyphens]{url}  
\usepackage{graphicx} 
\usepackage{natbib}  
\usepackage{caption} 
\usepackage{algorithm}
\usepackage{algorithmic}

\usepackage{xcolor}

\newcommand{\yh}[1]{\textcolor{black}{#1}}

\usepackage{newfloat}
\usepackage{listings}
\DeclareCaptionStyle{ruled}{labelfont=normalfont,labelsep=colon,strut=off} 
\floatstyle{ruled}
\newfloat{listing}{tb}{lst}{}
\floatname{listing}{Listing}
\usepackage{xspace}
\newcommand{\method}{{CMOOD}\xspace}

\usepackage{tikz}
\definecolor{darkgreen}{RGB}{60,179,113}

\definecolor{iccvblue}{rgb}{0.21,0.49,0.74}
\usepackage[utf8]{inputenc} 
\usepackage[T1]{fontenc}    
\usepackage{url}            
\usepackage{booktabs}       
\usepackage{amsfonts}       
\usepackage{nicefrac}       
\usepackage{microtype}      
\usepackage{xcolor}         
\usepackage{colortbl}

\usepackage{graphicx}
\usepackage{microtype}
\usepackage{graphicx}
\usepackage{amssymb}

\usepackage{mathtools}
\usepackage{amsthm}

\usepackage{array}
\usepackage{threeparttable}
\usepackage{lipsum}
\usepackage{arydshln}

\usepackage{amsfonts,amssymb}
\usepackage{multirow}
\usepackage{pifont}
\usepackage{booktabs}
\usepackage{siunitx} 

\usepackage{makecell}
\usepackage{dsfont}
\usepackage{amsmath}

\usepackage{acro}
\usepackage{algorithm}
\usepackage{algorithmic}
\usepackage{pifont}
\usepackage{siunitx}

\newcommand{\cmark}{\ding{51}}%
\newcommand{\xmark}{\ding{55}}%

\DeclareAcronym{NLP}{
  short = NLP,
  long = Natural Language Processing,
  tag = abbrev
}

\DeclareAcronym{VLM}{
  short = VLM,
  long  = Vision Language Model,
  tag = abbrev
}

\DeclareAcronym{LLM}{
  short = LLM,
  long  = Large Language Model,
  tag = abbrev
}

\DeclareAcronym{VLMs}{
  short = VLMs,
  long  = Vision Language Models,
  tag = abbrev
}

\DeclareAcronym{LLMs}{
  short = LLMs,
  long  = Large Language Models,
  tag = abbrev
}

\title{CMOOD: Concept-based Multi-label OOD Detection}
\author{
Zhendong Liu\textsuperscript{1,*},
Yi Nian\textsuperscript{1,*},
Yuehan Qin\textsuperscript{1,*},
Henry Peng Zou\textsuperscript{2},
Li Li\textsuperscript{1},
Xiyang Hu\textsuperscript{3}
}

\affiliations{
\textsuperscript{1}University of Southern California\\
\textsuperscript{2}University of Illinois Chicago\\
\textsuperscript{3}Arizona State University\\
lzd233a@gmail.com,
\{yinian, yuehanqi, li.li02\}@usc.edu,
pzou3@uic.edu,
xiyanghu@cmu.edu\\
\textit{\small $^*$Co-first authors}
}

\usepackage{bibentry}

\begin{document}

\maketitle

\begin{abstract}
How can models effectively detect out-of-distribution (OOD) samples in complex, multi-label settings without extensive retraining?
Existing OOD detection methods struggle to capture the intricate semantic relationships and label co-occurrences inherent in multi-label settings, often requiring large amounts of training data and failing to generalize to unseen label combinations.
While large language models have revolutionized zero-shot OOD detection, they primarily focus on single-label scenarios, leaving a critical gap in handling real-world tasks where samples can be associated with multiple interdependent labels.
To address these challenges, we introduce \method, a novel zero-shot multi-label OOD detection framework. \method leverages pre-trained vision-language models, enhancing them with a concept-based label expansion strategy and a new scoring function.
By enriching the semantic space with both positive and negative concepts for each label, our approach models complex label dependencies, precisely differentiating OOD samples without the need for additional training.
Extensive experiments demonstrate that our method significantly outperforms existing approaches, 
achieving approximately 95\% average AUROC on both VOC and COCO datasets, while maintaining robust performance across varying numbers of labels and different types of OOD samples. 
\begin{links}
\link{Code}{https://anonymous.4open.science/r/COOD-162F}
\end{links}

\end{abstract}
\section{Introduction}

As machine learning models become essential in a range of real-world applications, out-of-distribution (OOD) detection has gained increasing importance \cite{yang2024generalized}. 
OOD detection is particularly critical in fields such as autonomous driving \cite{elhafsi2023semantic}, medical diagnostics \cite{huang2024adapting}, and surveillance \cite{sultani2018real}, where detecting data that deviates from the training distribution is crucial to prevent safety risks or incorrect decisions \cite{hendrycks2019scaling}.
Thus, developing robust OOD detection techniques is key to ensuring model reliability in unpredictable environments.


\noindent \textbf{Large Models for OOD Detection (LM-OOD)}.
The rise of large models, particularly Vision-Language Models (VLMs) and MultiModal Large Language Models (MLLMs), has redefined the landscape of OOD detection \cite{xu2024large}. 
Traditional OOD detectors, such as Maximum Softmax Probability (MSP) and Mahalanobis distance-based techniques \cite{lee2018simple}, typically require extensive task-specific training on in-distribution (ID) data. In contrast, LLM-based methods leverage pre-trained knowledge, allowing for \textit{zero-shot} and \textit{few-shot} OOD detection.
Models like CLIP \cite{radford2021learning} exemplify this shift by achieving effective OOD detection with minimal task-specific training, relying on robust, pre-trained representations from multimodal datasets. 
Recent advancements, such as NegLabel \cite{jiang2024negative}, refine zero-shot OOD detection by introducing negative mining strategies to identify semantically meaningful OOD categories.
This highlights that carefully selected negative labels can significantly improve detection performance without retraining. 
LM-OOD methods offer distinct advantages over traditional approaches, including strong performance in limited data, high adaptability across various tasks, and improved computational efficiency through reduced dependence on task-specific data preparation and retraining \cite{ming2022delving, miyai2024generalized, xu2024large}.

 \begin{figure*}[t]
 \centering
          \includegraphics[width=\textwidth]{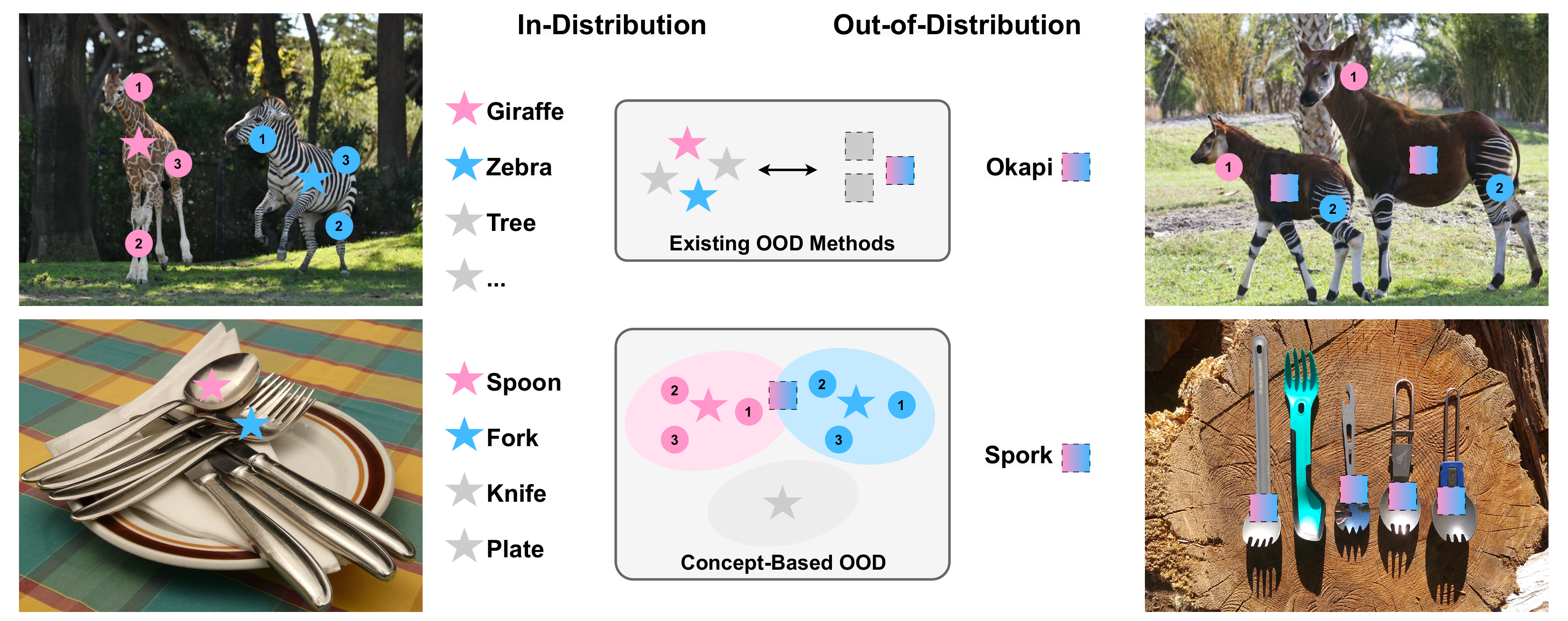} 
\caption{
Motivation for \method. Traditional methods struggle with complex multi-label cases. 
Our approach expands the label space with positive and negative concepts, enabling robust detection of complex OOD samples like ``Okapi" and ``Spork". 
}
    \label{fig:motivation}
    \end{figure*} 

\noindent
\textbf{Limitations in Current LM-OOD}.
Large models for OOD detection are primarily designed for single-label tasks, where each input corresponds to a single, definitive label \cite{fort2021exploring, ming2022delving}. 
However, many real-world applications are inherently multi-label in nature: medical imaging often requires identifying multiple coexisting conditions \cite{kermany2018identifying, huang2024adapting}.
Current state-of-the-art methods like NegLabel \cite{jiang2024negative} encounter limitations in these multi-label contexts, as seen in Fig.~\ref{fig:motivation}: 
First, the complexity of semantic similarity computations increases significantly with multiple concurrent labels compared to single labels. 
Second, these methods assume OOD samples are semantically distant from ID classes, which may not hold in multi-label settings where novel combinations of known concepts could form valid OOD cases \cite{wang2023clipn, nie2024out}. Additionally, these approaches struggle to effectively model complex label co-occurrence patterns and conditional dependencies, both crucial in multi-label contexts \cite{zhang2014review, yu2014large}.
These limitations highlight an essential gap in adapting LM-OOD methods for multi-label tasks, where it is necessary to differentiate between known and unknown label combinations within complex label spaces for practical application \cite{miyai2024generalized, yang2024generalized}.

\noindent
\noindent \textbf{Our Proposal: Extending LM-OOD to Multi-Label Settings}.
In this work, we introduce \method (concept-based OOD detection), the first multi-label OOD detection framework that leverages VLMs, specifically CLIP \cite{wang2023clipn}, in a zero-shot setting. 
\method addresses the challenges of multi-label OOD detection by incorporating a concept-based label expansion strategy. It enriches the base label set with two fine-grained concepts: positive concepts, which capture complex semantic details related to ID classes, and negative concepts, which, filtered by a similarity threshold, introduce semantically distant features to strengthen ID-OOD separation. \method embeds these expanded concept labels alongside the original base labels into a novel scoring function that accounts for both scenarios: OOD multi-label inputs that share more similarity with ID classes (e.g., Giraffe and Okapi) and those with less similarity to any ID class. This design enables precise detection of subtle distinctions between ID and OOD samples without requiring additional training.
Our technical coßntributions are summarized as follows:
\begin{itemize}
\item \textbf{Novel Multi-label OOD Detection}: We present a novel multi-label OOD detection framework based on the CLIP to achieve zero-shot detection without additional training.
\item \textbf{Concept-based Label Expansion}: We introduce a concept-based label expansion that leverages positive and negative concepts for precise discrimination of OOD samples and provide interpretability in multi-label tasks.
\item \textbf{Superior Performance and Efficiency}: Our method significantly outperforms existing approaches on standard benchmarks, with a high throughput of ~800 images per second on CLIP-B/16.
\end{itemize}

\section{Proposed \method Method}

\definecolor{stepOneColor}{RGB}{255, 225, 120} 
\definecolor{stepThreeColor}{RGB}{150, 180, 210} 

\begin{figure*}[t]
\centering
\includegraphics[width = 0.85 \textwidth]{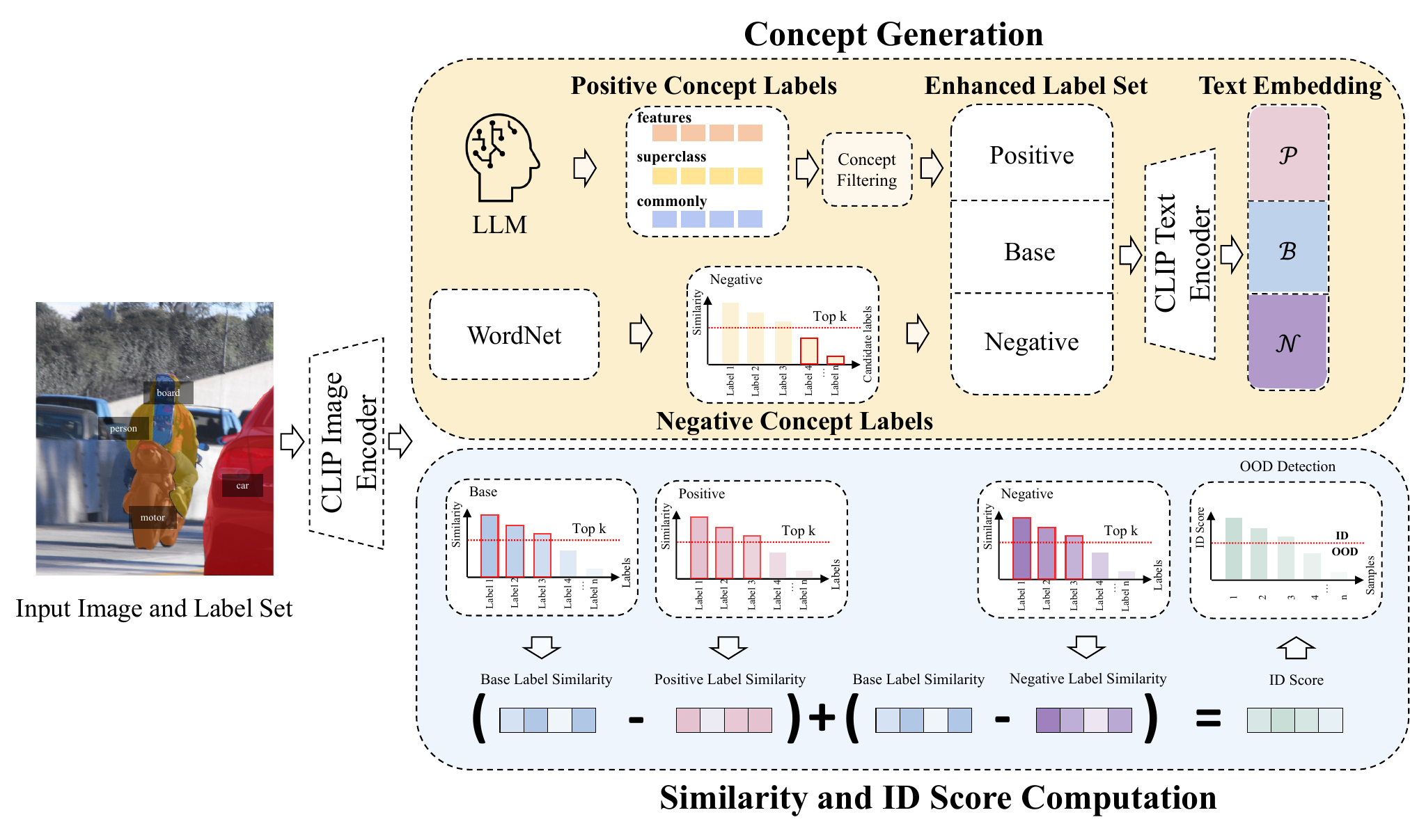} 
\caption{
Overview of \method
. The Concept Generation module
uses LLMs to expand base labels into positive ($\mathcal{P}$) and negative ($\mathcal{N}$) concept sets, enhancing the ID-OOD boundary. Positive concepts capture fine-grained, ID-aligned features, while negative concepts provide contrasting OOD-aligned features. 
The Similarity and ID Score Computation module 
encodes an input image and computes similarity scores.
An ID score based on top-$k$ similarities then classifies the image for ID/OOD.
}
\label{fig:method}
\end{figure*}
\subsection{Preliminaries on OOD Detection}

In multi-label OOD detection, we determine whether an input image $I \in \mathcal{I}$ is ID or OOD. Let 
$\mathcal{B} = \{\ell_1, \ell_2, \dots, \ell_{|\mathcal{B}|}\}$ denote the set of known ID classes, where each image $I$ can have multiple labels from $\mathcal{B}$.
We define an ID scoring function $S_{\text{ID}}: \mathcal{I} \rightarrow \mathbb{R}$
with the decision following criteria
\begin{equation}
    S_{\text{ID}}(I) \begin{cases}
        > \gamma, & \text{if } I \sim \mathcal{D}_{\text{ID}}, \\
        \leq \gamma, & \text{if } I \sim \mathcal{D}_{\text{OOD}},
    \end{cases}
\end{equation}
where $\gamma$ is a threshold determined through validation, and $\mathcal{D}_{\text{ID}}$, $\mathcal{D}_{\text{OOD}}$ are the ID and OOD distributions, respectively.

\subsubsection{Revisit VLM-based OOD Detection}
VLMs bring powerful multimodal capabilities for OOD detection, allowing more flexible, context-aware identification of OOD samples \cite{miyai2024generalized, xu2024large}. Large-scale pre-trained VLMs, such as CLIP \cite{radford2021learning} and GPT-4V \cite{zhang2023exploring}, enable models to process images and text prompts together, enhancing adaptability and precision across diverse visual and textual domains \cite{fort2021exploring}.

VLM-based OOD detection generally follows two primary approaches \cite{xu2024large}: \textit{Prompting-based Detection}, which directly prompts VLMs to respond with OOD indicators \cite{cao2023towards}, and \textit{Contrasting-based Detection}, which uses multimodal VLMs pre-trained with contrastive objectives to distinguish OOD samples \cite{ming2022delving}. 
This work focuses on the contrasting-based approach in multimodal contexts, as it is well-suited for enhancing OOD detection by amplifying distinctions between ID and OOD classes.
Methods like \texttt{NegLabel} \cite{jiang2024negative} and \texttt{NegPrompt} \cite{li2024learning} fall under this category. \texttt{NegLabel} enhances OOD detection by introducing negative labels to contrast with ID classes, while \texttt{NegPrompt} uses learned negative prompts to emphasize OOD differences by contrasting them with ID prompts.

\noindent
\textbf{Formal Definition}. The OOD detection score \( S(x) \) for a sample \( x \) is defined as:
\begin{small}
\begin{equation}
S_{OOD}(x) = \frac{\sum_{i \in Y} \exp\big(\text{sim}(h, e_i)\big)}{\sum_{i \in Y} \exp\big(\text{sim}(h, e_i)\big) + \sum_{j \in Y^-} \exp\big(\text{sim}(h, e_j^-)\big)},
\end{equation}
\end{small}
\noindent where \( h \) is the embedding of \( x \); \( e_i \) and \( e_j^- \) are embeddings for ID and negative representations (\texttt{NegPrompt} \cite{li2024learning}, \texttt{NegLabel}\cite{jiang2024negative}) from sets \( Y \) and \( Y^- \), respectively; \( \text{sim}(h, e) \) denotes the similarity (e.g., cosine similarity) between embeddings.
The numerator measures similarity to ID representations, while the denominator amplifies the detection signal with both ID and negative similarities.
Note that our definition is a little different: a smaller $S(x)$ means a larger probability of $x$ being an OOD sample.

\noindent
\textbf{Limitations in Current Approaches}.  
Current methods like NegLabel and NegPrompt assume that OOD classes are semantically distant from ID classes, which may not hold in real-world scenarios. OOD samples can closely resemble ID classes, particularly in fine-grained distinctions (e.g., similar dog breeds), limiting the effectiveness of a purely semantic distance-based approach.

Moreover, extending single-label OOD detection methods \cite{sun2021react, liu2020energy} to multi-label contexts presents additional challenges. Multi-label settings often involve significant semantic overlap between labels and co-occurring ID and OOD labels, complicating the extraction of effective negative labels or prompts. These issues highlight the need for advanced detection techniques to capture the complex dependencies and relationships inherent in multi-label OOD detection tasks.

\subsection{Overview of \method Method}
\label{subsec:overview}



In fact, it is not new to use LLM to assist in generating descriptions and then using CLIP models for classification tasks like DCLIP \cite{menon2022visual}. However, there are two problems with the application of this method: (1) the generated description only contains texts with similar semantic meanings, and the description of the text space is not comprehensive. (2) When considering the classification task, the OOD detection task does not bring improvement, and even brings negative benefits as show in Table \ref{table:main}. 

The \method framework improves OOD detection by refining the decision boundary between ID and OOD samples using fine-grained concepts. By introducing positive and negative concept sets, $\mathcal{P}$ and $\mathcal{N}$, \method leverages LLMs to add contextual information around ID samples, thereby enhancing sensitivity to OOD cases.

\method is built on two core components (see Fig.~\ref{fig:method}):
\begin{enumerate}
    \item \textbf{Concept Generation and Similarity Measure}:
    To distinguish ID from OOD samples, we generate positive concepts ($\mathcal{P}$) and negative concepts ($\mathcal{N}$) that are closely and distantly related to ID classes, respectively. 
    Positive concepts, mined using LLMs, capture domain-relevant features of ID samples at a fine-grained level, while negative concepts are chosen based on their semantic distance from ID classes to enhance contrast. 
    \item \textbf{Similarity and ID Score Computation}:
    For each input image \( I \), we compute similarity scores with these concept sets and the base label set $\mathcal{B}$, forming a robust semantic space for evaluating ID-OOD relationships.
    After this, we define an ID score \( S_{\text{ID}}(I) \) that aggregates the image’s alignment with base concepts ($\mathcal{B}$)
    and contrasts it with the positive and negative concepts ($\mathcal{P}$ and $\mathcal{N}$). When this score falls below a predefined threshold \( \gamma \), the image is classified as OOD. This scoring method sharpens the decision boundary by
    leveraging contrasts with positive and negative concepts
    giving more accurate OOD detection.
\end{enumerate}

The pseudocode in Algo.~\ref{alg:ood_detection} outlines the full procedure for \method.
To formalize the OOD detection decision, we compute an ID score and classify an image \( X \) as OOD if this score falls below a threshold \( \gamma \). Specifically, the decision function \( \tilde{Y} \) is defined as follows:
\begin{equation}
\begin{aligned}
   \tilde{Y} &= \mathbb{I}(S_{\text{ID}}(h, P, N) < \gamma), \quad \text{where} \quad h = f_{\text{img}}(X), \\
    P &= f_{\text{text}}(\text{prompt}(\mathcal{P})), \quad N = f_{\text{text}}(\text{prompt}(\mathcal{N})).
\end{aligned}
\end{equation}
Here, \( h \) is the image embedding for \( X \), obtained using an image encoder \( f_{\text{img}} \), while \( P \) and \( N \) are the positive and negative concept embeddings generated via text prompts for concept sets \( \mathcal{P} \) and \( \mathcal{N} \), respectively. The indicator function \( \mathbb{I} \) produces the final OOD classification.


\noindent
\textbf{Advantages.}  
The proposed \method approach is computationally efficient, leveraging pre-trained embeddings without requiring additional model training. By incorporating positive and negative concepts, the model gains enhanced semantic understanding, allowing for clearer differentiation between ID and OOD samples. The focus on top-$k$ similarity values makes the method robust to noise, ensuring stable performance across varied datasets and label sets.

\begin{algorithm}[!t]
\caption{Multi-Label OOD Detection with \method}
\label{alg:ood_detection}
\begin{algorithmic}[1]
\REQUIRE Image $I$, base label set $\mathcal{B}$, threshold $\gamma$, top-$k$ parameter $k$
\ENSURE Classification of $I$ as ID or OOD

\STATE \textbf{Concept Generation} (\S \ref{subsec:concept}): Generate positive concepts $\mathcal{P}$ and negative concepts $\mathcal{N}$ using an LLM.

\STATE \textbf{Embedding Computation}: Compute embeddings for $I$, as well as for each label in $\mathcal{B}$, $\mathcal{P}$, and $\mathcal{N}$.

\STATE \textbf{Top-$k$ Similarity Calculation} (\S \ref{subsec:score}): Calculate the top-$k$ mean similarity scores, $\mu_k(\mathcal{B}, I)$, $\mu_k(\mathcal{P}, I)$, and $\mu_k(\mathcal{N}, I)$.

\STATE \textbf{ID Score Computation} (\S \ref{subsec:score}): Using the top-$k$ similarities, compute the ID score $S_{\text{ID}}(I)$ according to Eq.~\eqref{eq:ood_score}.

\STATE \textbf{Decision}: \textbf{if} $S_{\text{ID}}(I) > \gamma$ \textbf{then} classify $I$ as ID; \textbf{else} classify $I$ as OOD.

\end{algorithmic}
\end{algorithm}



\subsection{Concept Generation}
\label{subsec:concept}
\noindent
\textbf{Motivation}.
The concept generation process in \method creates a rich semantic space that strengthens ID-OOD distinctions. By constructing a set of positive concepts, $\mathcal{P}$, aligned with ID samples, and a set of negative concepts, $\mathcal{N}$, that enhance the contrast with OOD samples, we can refine the decision boundary between ID and OOD. The sets $\mathcal{P}$ and $\mathcal{N}$ are complementary: $\mathcal{P}$ effectively captures multi-label inputs that share similarities with ID classes, while $\mathcal{N}$ captures multi-label inputs whose components are all dissimilar from ID samples. Together, these complementary concept sets enable \method to model both subtle distributional shifts and drastic deviations, leading to improved OOD detection across varying degrees of distribution shift.

\textbf{Positive Concept Mining $\mathcal{P}$.} 
To build a comprehensive set of positive concepts that accurately represents ID characteristics, we use a two-stage approach inspired by LF-CBM \cite{lf-cbm}. This process enriches each base label by capturing specific attributes that reinforce its identity within the ID class.
\begin{enumerate}
    \item \textbf{Concept Querying}: We prompt GPT-4 to generate concepts in three distinct contexts for each target object, using structured prompts tailored to elicit three categories: features (prompt$_F$), superclasses (prompt$_S$), and commonly associated items (prompt$_C$). This contextual querying ensures that each concept captures detailed, complex information, improving the clarity and consistency of the concept pool. 
    Each prompt is crafted to encourage concise, relevant responses without qualifiers, which enhances precision, detailed in Supplementary Material \S 3.1.
    \item \textbf{Concept Filtering}: After generating these candidate concepts, we apply filtering criteria to retain only the most distinctive and relevant features. 
    This step is essential to ensure that the concept set effectively captures the ID characteristics required to differentiate ID from OOD. Our filtering process, detailed in the Supplementary Material \S 3.1,
    refines the generated concepts into a cohesive and meaningful set of positive labels.
\end{enumerate}

The final positive concept set, $\mathcal{P}$, is constructed as: 
\begin{small}
\begin{equation} 
\mathcal{P} = \bigcup_{c_i \in \mathcal{B}} \left( \text{prompt}_{F}(c_i) \cup \text{prompt}_{S}(c_i) \cup \text{prompt}_{C}(c_i) \right), \end{equation} 
\end{small}
where $\mathcal{B}$ denotes the set of ID labels, and each $c_i \in \mathcal{B}$ is queried to generate its positive concept labels.

\noindent
\textbf{Negative Concept Mining $\mathcal{N}$.}
To reinforce the distinction between ID and OOD samples, we develop a set of negative concepts, $\mathcal{N}$, using a process called NegMining \cite{jiang2024negative}. This approach helps us identify concepts that are semantically distant from ID labels, forming a contrasting boundary that enhances OOD detection.

Starting with a large collection of words from a lexical database like WordNet, we create a candidate label space \( \mathcal{N}^c = \{n_1, n_2, \dots, n_C\} \). For each candidate label \( \tilde{n}_i \) in this set, we calculate a similarity score based on its cosine similarity with the entire ID label set, ensuring that the chosen negative concepts have minimal alignment with ID labels.
\begin{itemize}
    \item For each ID label \( l \in \mathcal{B} \), we obtain its text embedding \( e_l \) using a text encoder \( f^{\text{text}} \).
    \item For each candidate \( \tilde{n}_i \in \mathcal{N}^c \), we compute its embedding:
    \begin{equation}
        \tilde{e}_i = f^{\text{text}}(\text{prompt}(\tilde{n}_i)).
    \end{equation}
\end{itemize}
We select the top candidates with the smallest similarity scores relative to the ID label set, forming a set of negative labels that are maximally distant from ID concepts. 

The final negative concept set is defined as:
\begin{equation}
\begin{aligned}
    \mathcal{N} &= \left\{ n \mid \text{Sim}(\tilde{n}_i, \mathcal{B}) < \tau_{i} \right\}, \\
    \tau_i &= \text{percentile}_\eta \left( \{\text{Sim}(\tilde{e}_i, e_l)\}_{l \in \mathcal{B}} \right),
\end{aligned}
\end{equation}
where \( \text{Sim}(\tilde{n}_i, \mathcal{B}) \) represents the similarity between candidate \( \tilde{n}_i \) and the ID set \( \mathcal{B} \), and \( \tau_i \) is the \( \eta \)-th percentile of similarity scores, providing robustness against outliers.

By expanding the base label set $\mathcal{B}$ to include positive concepts $\mathcal{P}$ and contrasting them with negative concepts $\mathcal{N}$, we create an enriched semantic space. 
This structure allows for more precise discrimination between ID and OOD samples, capturing both the core characteristics of the ID classes and
\yh{intra-class and inter-class similarity}
effectively.

\begin{figure}[t]
 \centering
          \includegraphics[width=0.5 \textwidth]
          {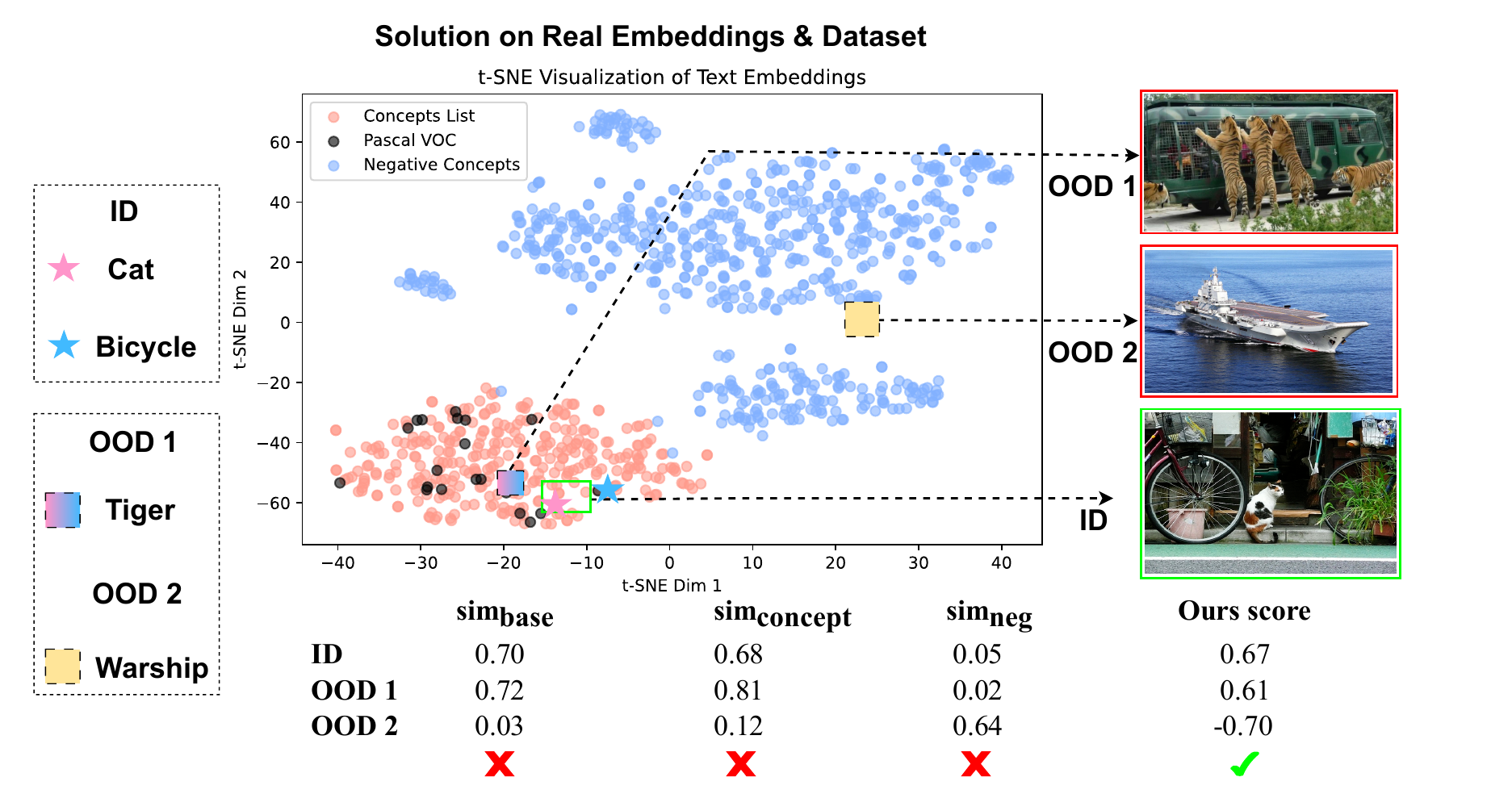} 
\caption{
t-SNE figure of label and concept text embeddings, along with corresponding image examples. In scenarios with multiple labels and objects, it is difficult to model the OOD detection problem using a single similarity measure. Instead, the COOD method is employed to address this issue.
}
 \label{fig:tsne}
\end{figure}

\subsection{Similarity and ID Score Computation}
\label{subsec:score}

After generating our positive and negative concepts, \method computes an ID score for each input image \( I \), based on its similarity with the positive concept set \( \mathcal{P} \), negative concept set \( \mathcal{N} \), and base labels \( \mathcal{B} \). This score helps determine if the image \yh{exhibits more}
ID or OOD characteristics. As shown in Figure \ref{fig:tsne}, the positive concepts are
distributed around the base labels, while the negative concepts are located at a relatively distant position in the text embedding space, showing a significant distinction. Therefore,
\yh{to capture this distinction, we model the problem by adjusting the base label-image similarity through two components: }
(1)
\yh{Subtracting the intra-class similarity between nearby positive concepts and the image, which captures within-distribution variations. Positive concepts are useful when identifying multi-label OOD inputs where it has some similarities with ID class like Giraffe and Okapi in Fig.~\ref{fig:motivation}.} (2)
\yh{Subtracting the inter-class similarity between distant negative concepts and the image, which captures cross-distribution relationships. This is to penalize the score when multi-label OOD input is not relevant to ID class.}

\noindent
\textbf{ID Score Definition.}  
The ID score \( S_{\text{ID}}(I) \) for an image \( I \) is:
\begin{equation}
S_{\text{ID}}(I) = \underbracket{\left[ \mu_k(\mathcal{B}, I) - \mu_k(\mathcal{P}, I) \right]}_{S_{\text{A}}} + \underbracket{\left[ \mu_k(\mathcal{B}, I) - \mu_k(\mathcal{N}, I) \right]}_{S_{\text{B}}}
\label{eq:ood_score}
\end{equation}
where \( \mathcal{B} \) are the base labels, \( \mathcal{P} \) are the positive concepts, and \( \mathcal{N} \) are the negative concepts.

In general, when an out-of-distribution (OOD) input has some similarity to an in-distribution (ID) class, we penalize $S_A$, whereas when an OOD input has no similarity to any ID class, we penalize $S_B$. However, for ID inputs, both $S_A$ and $S_B$ are larger. Consequently, ID inputs receive higher scores, while OOD inputs receive lower scores. Please refer to the theoretical bounds for details in Supplementary Material \S 2 and Fig. 1 that validate the scoring function’s effectiveness.

\noindent
\textbf{Top-\(k\) Mean Similarity Calculation.}  
The term \( \mu_k(\mathcal{S}, I) \) in Eq. (\ref{eq:ood_score}) represents the top-\(k\) mean similarity score between the image \( I \) and the label set \( \mathcal{S} \), calculated as:
\begin{equation}
\mu_k(\mathcal{S}, I) = \frac{1}{k} \sum_{i=1}^k \text{sim}(h, e_i)
\end{equation}
where \( h \) is the embedding of image \( I \), and \( e_i \) are the embeddings of the top-\(k\) closest elements in set \( \mathcal{S} \). Here, \( \mathcal{S} \) can be any of \(\mathcal{B}\) (base labels), \(\mathcal{P}\) (positive concepts), or \(\mathcal{N}\) (negative concepts). The similarity function \( \text{sim}(h, e_i) \) (e.g., cosine similarity) captures the alignment between the image embedding \( h \) and each of the concept embeddings \( e_i \).

This top-\(k\) similarity measure emphasizes the most relevant matches between the image and the concepts in each set, minimizing the influence of less relevant or noisy features and enhancing robustness in scoring.


\noindent
\textbf{Interpretation of the ID Score.}  
The scoring function \( S_{\text{ID}}(I) \) captures the degree of alignment of an image with\yh{in} ID or \yh{with} OOD characteristics. For ID samples, we anticipate high alignment with \yh{base labels \( \mathcal{B} \)}
, yielding a high \( S_{\text{ID}}(I) \). Conversely, OOD samples will likely align more closely with negative concepts \( \mathcal{N} \), resulting in a lower ID score. By focusing on the top-\(k\) similarities, this approach enhances robustness to noise and emphasizes the most semantically relevant features.

This mechanism refines the boundary between ID and OOD samples by leveraging positive and negative contrasts \yh{to base labels}, resulting in more accurate and reliable OOD detection.

\begin{table*}[h]
\centering
\small
\caption{
OOD Detection Results on Various Datasets. Our ViT-based CLIP model achieves strong performance with the lowest FPR95 and highest AUROC across most datasets, outperforming standard OOD methods. The ResNet-based CLIP variant also performs competitively. 
}

\label{table:main}

\scalebox{0.75}{
\small
\begin{tabular}{lcccccccccc}
\toprule
\textbf{ID Dataset} & \multicolumn{4}{c}{\textbf{Pascal VOC}} & \multicolumn{4}{c}{\textbf{COCO}} & \multicolumn{2}{c}{\textbf{ImageNet}} \\
\cmidrule(lr){2-5} \cmidrule(lr){6-9} \cmidrule(lr){10-11}
\textbf{OOD Dataset} & \multicolumn{2}{c}{ImageNet22k} & \multicolumn{2}{c}{Textures} & \multicolumn{2}{c}{ImageNet22k} & \multicolumn{2}{c}{Textures} & \multicolumn{2}{c}{Textures} \\
\cmidrule(lr){2-3} \cmidrule(lr){4-5} \cmidrule(lr){6-7} \cmidrule(lr){8-9} \cmidrule(lr){10-11}
\textbf{Method} & \textbf{FPR95}$\downarrow$ & \textbf{AUROC}$\uparrow$ & \textbf{FPR95}$\downarrow$ & \textbf{AUROC}$\uparrow$ & \textbf{FPR95}$\downarrow$ & \textbf{AUROC}$\uparrow$ & \textbf{FPR95}$\downarrow$ & \textbf{AUROC}$\uparrow$ & \textbf{FPR95}$\downarrow$ & \textbf{AUROC}$\uparrow$ \\
\midrule
\multicolumn{11}{l}{\textbf{ResNet-based Multi-label Classifier}} \\ 
MaxLogit{*} \cite{hendrycks2019benchmark} & 36.32 & 91.04 & 12.36 & 96.22 & 44.47 & 87.13 & 19.83 & 95.31 & 57.09 & 86.71 \\
MSP{*} \cite{hendrycks2016baseline} & 69.85 & 78.24 & 41.81 & 89.76 & 82.15 & 67.47 & 65.21 & 81.88 & 68.00 & 79.61 \\
ODIN{*} \cite{liang2017enhancing} & 36.32 & 91.04 & 12.36 & 96.22 & 54.51 & 84.92 & 33.15 & 90.71 & 50.23 & 85.62 \\
(Joint)-Energy{*} \cite{liu2020energy, wang2021can} & 31.96 & 92.32 & 10.87 & 96.78 & 41.81 & 90.30 & 17.72 & 96.07 & 53.72 & 85.99 \\
\textbf{Ours (ResNet-based CLIP)} & 25.28 & 93.32 & \textbf{8.76} & \textbf{97.79} & 26.83 & 93.14 & \textbf{10.53} & \textbf{97.17} & 43.72 & 91.68 \\ 
\midrule
\multicolumn{11}{l}{\textbf{ViT-based CLIP}} \\
MSP{\dag} \cite{hendrycks2016baseline} & 86.35 & 75.42 & 64.74 & 85.46 & 59.27 & 87.30 & 45.57 & 90.35 & 64.96 & 78.16\\
(Joint)-Energy{\dag} \cite{liu2020energy, wang2021can} & 81.93 & 76.59 & 90.18 & 78.23 & 65.39 & 85.20 & 76.01 & 82.40 & 51.18 & 88.09 \\
DCLIP{*}\cite{menon2022visual} & 96.96 & 61.59 & 97.94 & 65.26 & 89.50 & 71.43 &  72.77 & 83.34 & 91.08 & 70.75 \\
MCM{\dag *} \cite{ming2022delving} & 73.81 & 80.37 & 53.67 & 88.52 & 63.34 & 86.10 & 49.22 & 89.11 & 57.77 & 86.11\\
GL-MCM{\dag *} \cite{miyai2024generalized} & 72.98 & 81.76 & 64.74 & 86.96 & 48.96 & 88.50 & 45.06 & 89.70 & 57.41 & 83.73\\
NegLabel{*}\cite{jiang2024negative} & 35.83 & 91.18 & 43.14 & 89.72 & 33.24 & 90.19 & 47.33 & 85.10 & 43.56 & 90.22 \\
MCM+SeTAR{\dag *} \cite{li2024setar} & 48.25 & 92.08 & 40.44 & 93.58 & 73.55 & 80.43 & 47.33 & 89.58 & 55.83 & 86.58 \\
GL-MCM+SeTAR{\dag *} \cite{miyai2024generalized, li2024setar} & 31.47 & 94.31 & 20.35 & 96.36 & 65.30 & 81.38 & 42.05 & 89.81 & 54.17 & 84.59 \\
CLIPN{*}\cite{wang2023clipn} & 64.78  & 82.46 & 37.42 & 92.63 & 44.63  & 89.21 & 25.37 & 94.08 & 40.83 & 90.93 \\
CLIPScope{*}\cite{fu2024clipscope}  & 70.86  & 85.51 & 77.42  & 83.04 & 55.90  & 87.63 & 73.63  & 78.71 & \textbf{38.37} & \textbf{91.41} \\
\textbf{Ours (ViT-based CLIP)} & \textbf{23.87} & \textbf{94.32} & 21.58 & 95.14 & \textbf{20.37} & \textbf{95.07} & 21.63 & 94.53 & 39.41 & 91.10 \\
\bottomrule
\end{tabular}
}
\end{table*}

\section{Experimental Results}
        
\subsection{Experimental Setup}

\textbf{Datasets and Settings.} 
We evaluate our proposed multi-label OOD detection method on widely used datasets. For ID datasets, we use Pascal VOC \cite{everingham2010pascal}, MS-COCO \cite{coco}, and Objects365 \cite{shao2019objects365}. 
For OOD datasets, we use Textures \cite{cimpoi14describing} and Filtered ImageNet22K \cite{hendrycks2019benchmark}. These datasets provide diverse categories and rich multi-label annotations, making them ideal for assessing OOD detection in complex real-world scenarios. To quantify the OOD detection performance, we report two standard metrics, \textbf{FPR@95} and \textbf{AUROC}.
See
Supplementary Material \S 4.1
for more datasets and metrics details.


\noindent
\textbf{Implementation Details.} 
Our method is implemented using PyTorch and evaluated on   NVIDIA 3090 GPU. As a zero-shot approach, our method leverages pre-computed concept embeddings and does not require any additional training. This results in a computationally efficient solution, with an average inference time of approximately 800 images per second. 
This efficient design allows for parallelized OOD detection alongside zero-shot classification tasks, making it suitable for deployment in real-time applications.
For reproducibility, all hyperparameters and settings follow standard configurations, and further implementation details are provided in Supplementary Material \S 4.1.


\begin{figure*}[t]
\centering
\includegraphics[width = \textwidth]{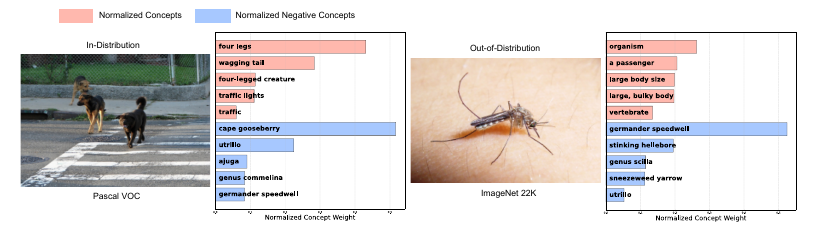} 
\caption{Analysis of \method on ID (left) and OOD (right) examples. For ID sample (dogs), positive concepts (e.g.,``four-legged creature") receive high weights, confirming ID alignment. 
For OOD samples (mosquito), negative concepts dominate.}
\label{fig:explain}
\end{figure*}

\subsection{Experimental Results and Analysis}

\textbf{OOD Detection Performance Comparison.}
Table \ref{table:main} presents a comparison of the OOD detection performance of \method against several baseline approaches. Where \textbf{*} represents our reproduction of the missing results from the original paper, and \textbf{\dag} means that we used the values from the original paper, and \textbf{\dag *} represents the best value we take from the replicated results and the reported results in the original paper. The gray boxes represent our approach. In addition, \textbf{(Joint)-Energy} denotes the better of the JointEnergy \citep{wang2021can} method and the Energy \citep{liu2020energy} method. \textbf{Bold} values indicate the best results.
\textit{Our method demonstrates strong OOD detection performance across both ResNet-based and ViT-based architectures}.
Specifically, our ResNet-based CLIP achieves superior results, particularly on the Pascal VOC ID dataset and Textures OOD dataset, with FPR95 of 8.76\% and AUROC of 97.79\%, outperforming all baseline methods. These suggest that the ResNet-based architecture may be particularly effective in handling texture-oriented datasets.

On the ViT-based CLIP, our method achieves the lowest FPR95 of 23.87\% on the Pascal VOC dataset and outperforms competing methods on the COCO dataset, achieving an impressive AUROC of 95.07\%. The ViT-based approach performs exceptionally well across most datasets, indicating that the ViT architecture enhances OOD detection across diverse image distributions, while still maintaining strong robustness on texture-based datasets.
For more OOD dataset like NINCO \cite{bitterwolf2023ninco} and SSB \cite{vaze2021open}, please see Table \ref{tab:moreood}. Our \method also demonstrates strong competitiveness.

\begin{table}[htbp]
    \centering
    \caption{ AUROC / FPR@95 for more OOD datasets.}
    \label{tab:moreood}
    \scalebox{0.75}{
    \begin{tabular}{llccc}
        \toprule
        ID Dataset & Method & {NINCO} & {SSB-easy} & {SSB-hard} \\
        \midrule
        COCO2017 & \method & 85.89 / 40.58 & 82.71 / 54.30 & 87.72 / 45.19 \\
        Pascal VOC & \method & 88.79 / 42.17 & 88.75 / 43.71 & 72.70 / 59.62 \\
        \midrule
     \multirow{4}{*}{ImageNet}    & MSP & 69.32 / 81.09 & 80.06 / 80.79 & 55.83 / 93.27 \\
     & MCM & 68.80 / 83.39 & 80.20 / 78.15 & 55.83 / 93.27 \\
          & NegLabel & 72.96 / 72.81 & 71.22 / 80.13 & 55.40 / 88.94 \\
         & \method & \textbf{77.40} / \textbf{69.91} & \textbf{84.04} / \textbf{69.54} & \textbf{64.67} / \textbf{81.25}\\

        \bottomrule
    \end{tabular}}
\end{table}

Importantly, our approach operates as a \textit{zero-shot} method, requiring no additional training or complex parameter tuning. This makes it highly efficient with low computational overhead. This contrasts with other multi-label classifiers requires training. For method without training, such as SeTAR \cite{li2024setar}, which requires extensive parameter search time, impacting practical deployment. Overall, our method provides an efficient and effective zero-shot solution for OOD detection, yielding state-of-the-art performance with minimal overhead on both architectures.


\subsection{Visual Explainability of \method Components}

Fig. \ref{fig:tsne} and \ref{fig:explain} present a detailed analysis of \method through t-SNE visualization \cite{van2008visualizing} of text embeddings and qualitative examples. 
\method's explainability
is a key feature, allowing us to
understand how specific concepts contribute to OOD detection.

\noindent
\textbf{t-SNE Visualization.} 
The t-SNE plot on the Fig. \ref{fig:tsne} visualizes the embeddings of concepts from our model, showing a clear separation between ID labels (Pascal VOC, shown in red) and OOD concepts. Notably, the embeddings for positive concepts $\mathcal{P}$ (shown in red) form distinct clusters separate from negative concepts $\mathcal{N}$ (shown in blue)\yh{, while the black points represent the base labels $\mathcal{B}$}. This clustering reflects the model's ability to encode meaningful semantic information that distinguishes ID from OOD samples. The clustering of positive concepts around ID points and the scattering of negative concepts in more distant regions confirm that the model effectively captures relationships that contribute to its decision-making process.

\noindent
\textbf{Qualitative Examples.} 
Fig. \ref{fig:explain} shows example images
with corresponding bar plots, illustrating the concept weights associated with each image. In each bar plot, positive concepts are marked in red, while negative concepts are marked in blue. This visual representation offers insight into which semantic features are influencing the OOD score, making it easier to interpret why certain samples are classified as ID or OOD.


\subsection{Ablation Studies and Additional Analysis}
To assess the contributions of different components in our approach, we conduct an ablation study with variations in the parameters
$S_{\text{A}}$ and $S_{\text{B}}$ on the Pascal VOC and COCO datasets, as shown in Table \ref{table:ablation}. For more experiments on different model architectures and OOD score functions, please refer to
Supplementary Material \S 4.
Recall that we defined the score function in Eq. (\ref{eq:ood_score}).
Enabling both $S_{\text{A}}$ and $S_{\text{B}}$ yields the best performance across metrics. For example, on Pascal VOC, activating both scores reduces FPR95 from 36.12\% to 24.78\% and boosts AUROC from 91.72\% to 94.27\%. This improvement indicates that $S_{\text{A}}$ and $S_{\text{B}}$ capture complementary aspects of the data, thereby enhancing the model's ability to differentiate between ID and OOD samples effectively.

\begin{table}[h]
\centering
\caption{Ablations on Pascal VOC and COCO datasets.
Using both $S_{\text{A}}$ and $S_{\text{B}}$ together yields the best results.
}
\label{table:ablation}
\scalebox{0.9}{
\begin{tabular}{cc|cc|cc}
\toprule
$S_{\text{A}}$ & $S_{\text{B}}$ & \multicolumn{2}{c|}{\textbf{Pascal VOC ID}} & \multicolumn{2}{c}{\textbf{COCO ID}} \\
\cmidrule{3-6}
& & \textbf{FPR95} & \textbf{AUROC} & \textbf{FPR95} & \textbf{AUROC} \\
\midrule
 \xmark & \cmark & 36.12 & 91.72 & 48.87 & 89.49 \\
\midrule
 \cmark & \xmark & 36.43 & 91.73 & 33.70 & 90.09 \\
\midrule
 \cmark & \cmark & 24.78 & 94.27 & 29.08 & 92.39 \\
\bottomrule
\end{tabular}
}
\end{table}
\section{Related Work (Extended Ver. in Supp. \S 1)}
CLIP-based OOD detection methods include MCM \citep{ming2022delving}, which uses textual embeddings as concept prototypes, NegLabel \citep{jiang2024negative}, which introduces semantically distant negative labels, and SeTAR \citep{li2024setar}, which applies selective low-rank approximation. However, these methods primarily focus on single-label scenarios.

Multi-label OOD detection has been explored by JointEnergy \citep{wang2021can} and YolOOD \citep{zolfi2024yolood}, but these don't utilize CLIP architectures. Our work is the first to comprehensively address zero-shot multi-label OOD detection using CLIP models.

\section{Conclusion, Limitations, and Broader Impact}
We introduced \method, a novel framework that leverages concept-based reasoning for zero-shot Out-of-Distribution (OOD) detection in complex multi-label environments. Our approach sets a new state of the art, achieving over 95\% average AUROC on VOC and COCO by effectively distinguishing In-Distribution data from challenging OOD datasets like ImageNet and Texture. Through comprehensive experiments, we also demonstrated the method's high degree of explainability and robustness. This work validates that dissecting scenes into fine-grained concepts is a powerful and efficient paradigm for multi-label OOD detection.


Our work builds upon the powerful capabilities of large-scale language models. Like all methods leveraging such foundation models, the performance of \method is correlated with the quality of the embedding space and the richness of the provided concepts. This connection does not represent a limitation of our framework, but rather highlights a frontier for future innovation. We identify two promising research trajectories:
(1) Learning concept vocabularies directly from data or external knowledge graphs, moving towards a more adaptive and scalable system beyond concepts from language models.
(2) Adapting our method to specialized domains. In medical image processing, for instance, this could enable the detection of rare diseases as OOD.

\method enhances AI reliability in safety-critical applications, such as healthcare and autonomous systems, by improving robustness against unknown data. This directly mitigates the risks of false detections, which can have severe consequences, thus contributing to safer performance in dynamic, high-stakes environments.


\bibliography{aaai25}

\clearpage

\end{document}